\documentclass{article}
\usepackage{cite}
\usepackage{spconf,amsmath,graphicx}
\usepackage{float}
\usepackage{chngcntr} %
\counterwithout{figure}
{section}
\usepackage{graphicx}
\usepackage{textcomp}
\usepackage{url}
\usepackage{hyperref}
\usepackage{slashbox}

\DeclareMathAlphabet      {\mathbfit}{OML}{cmm}{b}{it}
\title{Decoding Human Activities: Analyzing Wearable Accelerometer and Gyroscope Data for Activity Recognition}
\name{Utsab Saha$^{1}$, Sawradip Saha$^{2}$, Tahmid Kabir$^{1}$, Shaikh Anowarul Fattah$^{1}$, Mohammad Saquib$^{3}$}
\address{$^1$ Dept. of EEE, BUET, Bangladesh\\
$^2$ Dept. of ME, BUET, Bangladesh\\
$^3$Dept. of EE, UT Dallas, Texas, USA\\
 }
\begin{document}
%
\maketitle
\begin{abstract}

A person's movement or relative positioning can be effectively captured by different types of sensors and corresponding sensor output can be utilized in various manipulative techniques for the classification of different human activities. This letter proposes an effective scheme for human activity recognition, which introduces two unique approaches within a multi-structural architecture, named FusionActNet. The first approach aims to capture the static and dynamic behavior of a particular action by using two dedicated residual networks and the second approach facilitates the final decision-making process by introducing a guidance module. A two-stage training process is designed where at the first stage, residual networks are pre-trained separately by using static (where the human body is immobile) and dynamic (involving movement of the human body) data. In the next stage, the guidance module along with the pre-trained static/dynamic models are used to train the given sensor data. Here the guidance module learns to emphasize the most relevant prediction vector obtained from the static/dynamic models, which helps to effectively classify different human activities. The proposed scheme is evaluated using two benchmark datasets and compared with state-of-the-art methods. The results clearly demonstrate that our method outperforms existing approaches in terms of accuracy, precision, recall, and F1 score, achieving 97.35\% and 95.35\% accuracy on the UCI HAR and Motion-Sense datasets, respectively which highlights both the effectiveness and stability of the proposed scheme. 

\end{abstract}
\section{Introduction}
\label{sec:intro}
{A}{ctivity} recognition using wearable sensors has been a trending topic of research for its widespread applicability in diverse sectors ranging from healthcare services to military applications. Modern mobile devices provide abundant sensor data, which are valuable for applications like activity recognition. Various types of sensor data along with image and video data have been employed for recognizing human activity\cite{jalal2017robust}. In this letter, the time series wearable sensor data (e.g. 3-axis accelerometer and gyroscope) are mainly focused on, as they are pretty easy to obtain and can be used to recognize human activity from distant data which are very small in volume and easy to share through internet. In numerous literature, support vector machine (SVM), and k\textsuperscript{th} nearest neighbor (KNN) classifiers are the most popular ML-based training method for human activity recognition. Furthermore, multiple-stage training through LSTM-CNN-LSTM, ANN, 
CNN-LSTM \cite{rl16}, 
and several deep learning frameworks \cite{10288050}, \cite{10238776} are also suggested by some recent literatures \cite{semwal2022pattern}, \cite{jain2022deep}, \cite{bijalwan2022wearable}. However, augmented-signal features and a hierarchical recognizer are combined to get a highly functional trained structure reported in \cite{khan2010triaxial}. Reviewing existing literature it can be concluded that achieving a high level of accuracy in classifying human activities with raw 1-D time domain data from accelerometers and gyroscope sensors is challenging due to the inability of shallow networks to extract meaningful patterns, especially for similar activities like walking, lying, and sitting. These activities often exhibit substantial data overlap, making accurate prediction difficult for a single network.

To tackle the problem, this letter proposes a scheme for human activity recognition that aims to classify activities by capturing their static and dynamic behavior, incorporating a guidance mechanism.
 The key contributions of the paper are-
\begin{itemize}
  \item Introduction of a scheme named FusionActNet that aims to capture the static and dynamic behavior of a particular action by using two separate dedicated networks consisting of residual blocks. 
   \item Incorporation of a guidance mechanism to facilitate the final decision-making process, that relies on efficient training of static/dynamic models. In the guidance network, depthwise separable CNN blocks are used for emphasizing the prediction vectors generated by the pre-trained models. 
   \end{itemize}
The proposed scheme can help to improve robot-based assistance by enabling better understanding and response to human activities and enhancing personalized support and monitoring. Additionally, it can be used in surveillance systems by accurately identifying and tracking activities, enabling effective responses in dynamic scenarios.
The remaining sections of the letter are organized in the following order. In Section \ref{sec:Dset}, a brief description of the datasets used in this work, is presented. In Section \ref{sec:propose}, the problems addressed in this research work and the detailed architecture of the proposed network are presented. Section \ref{sec:res} depicts the experiment and result whereas section \ref{sec:con} comprises the conclusion of the paper.
\\
\section{Dataset Description}
\label{sec:Dset}
In this work, a publicly accessible dataset named UCI HAR is used \cite{anguita2013public}. It involved 30 volunteers aged 19 to 48, wearing a Samsung Galaxy S II smartphone on their waist while performing six activities which are as follows: Walking, Walking upstairs, Walking downstairs, Sitting, Standing, and Lying. Among these, Sitting, Standing, and Lying have been considered as \textit{Static} activities, and Walking, Walking upstairs, and Walking downstairs have been considered as \textit{Dynamic} activities for this work. The smartphone's accelerometers and gyroscopes recorded 3-axial linear acceleration and angular velocity at 50Hz. The dataset was split into 70\% (training and validation) and 30\% ratio (testing data). Pre-processing included noise filtering, fixed-width sliding windowing, and separating gravitational and body motion components. The dataset comprised 7352 training observations and 2947 test observations from 30 subjects for model evaluation.

Another dataset used in this work is the Motion-Sense dataset, which consists of 15 trials involving 24 individuals using an iPhone 6s placed in their trouser pockets to record accelerometer and gyroscope data \cite{malekzadeh2018protecting}. It includes 12 features, such as attitude, gravity, rotation rate, and user acceleration, and covers activities like Sitting, Standing, Walking, Walking Downstairs, Walking Upstairs, and Jogging. This dataset contains 2 Static classes and 4 Dynamic classes. The same split ratio is also incorporated for the experimentation (70\% for training and validation, and 30\% for test).

\begin{figure}[ht]
    \centering    \includegraphics[width=0.48\textwidth, height=0.16\textheight]{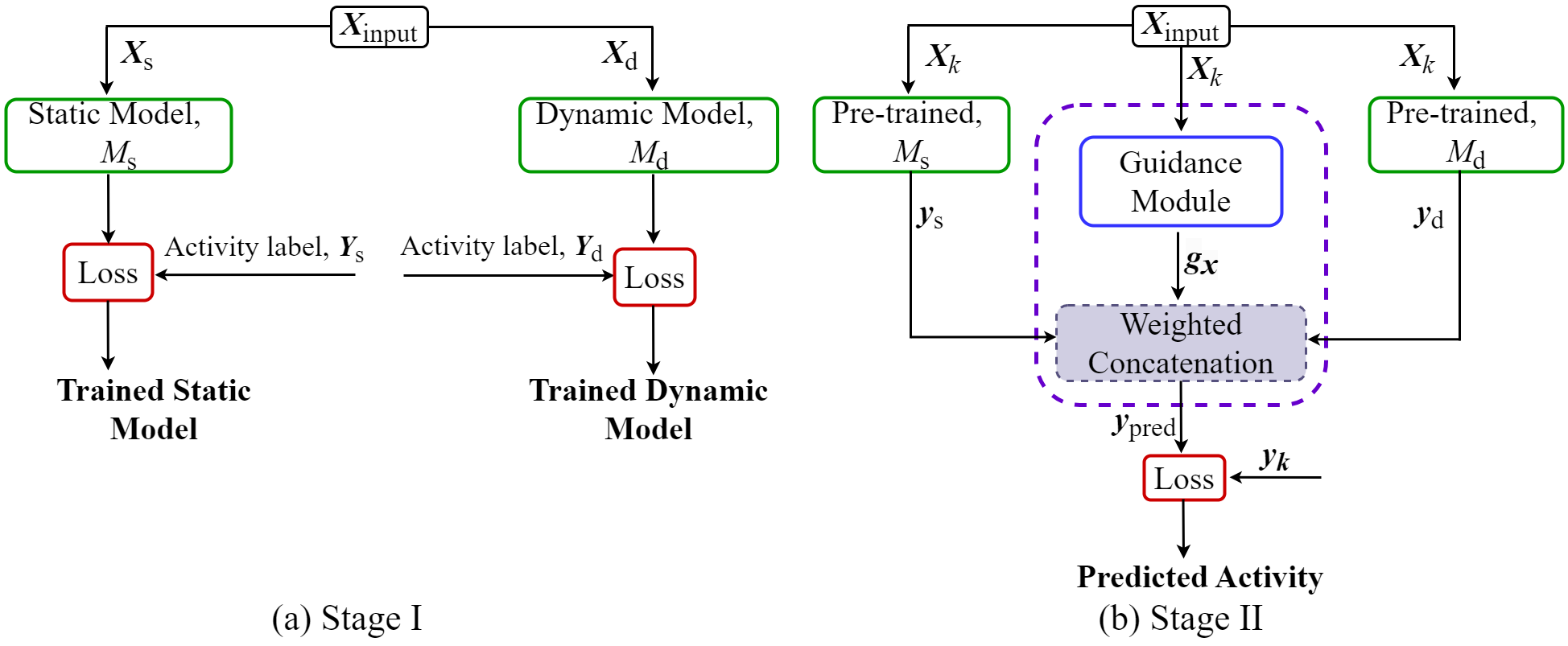}
    \caption{Simplified architecture of proposed FusionActNet. (a) Stage I (b) Stage II}
    \label{fig:propose2}
\end{figure}
\section{Methodology}
Within the domain of analyzing mixed-class activities, features retrieved from gyroscope and accelerometer signals become strongly correlated, which poses a considerable challenge in classifying these activities using a single deep learning network. To address this issue, the concept of categorizing human activities into two superclasses: Static and Dynamic, is introduced here. At first, these two superclass data are trained by two separate models, allowing each model to be trained with a specific type of data. In the second training stage, the two pre-trained models provide predictions for specific input data, which are then combined using guidance-based weighted concatenation to ensure proper superclass assignment and improve prediction accuracy.

\label{sec:propose}
\subsection{Problem Formulation}
In mathematical terms, a given data set of accelerometer and gyroscope is denoted by $S = {(\mathbfit{X}_1, \mathbfit{Y}_1), (\mathbfit{X}_2, \mathbfit{Y}_2), ... (\mathbfit{X}_n, \mathbfit{Y}_n )}$ where $n$ is the number of samples, $\mathbfit{X}_i$ denotes the $i^{th}$ inertial signal data and $\mathbfit{Y}_i$ represents the $i^{th}$ activity label. As mentioned before, the dataset is divided into two activity superclasses - static and dynamic data, each comprising the samples of all the static and dynamic activity classes, respectively. Let us denote static data as $S_\text{s} = {(\mathbfit{X}_\text{s1}, \mathbfit{Y}_\text{s1}) ... (\mathbfit{X}_{\text{s}p} , \mathbfit{Y}_{\text{s}p})}$ and dynamic data as $S_\text{d} = {(\mathbfit{X}_{\text{d}1}, \mathbfit{Y}_{\text{d}1}) ... (\mathbfit{X}_{\text{d}q} , \mathbfit{Y}_{\text{d}q})}$, where $p+q=n$. At the initial training stage, a static model $M_\text{s}$ is trained using $S_\text{s}$ and a dynamic model  $M_\text{d}$ is trained using $S_\text{d}$ as shown in Fig. \ref{fig:propose2}. After the first stage training process, two expert models are obtained that are specialized at classifying static and dynamic activity data respectively.

\begin{figure}[ht]
    \centering
    \includegraphics[width=0.44\textwidth, height=0.37\textheight]
    {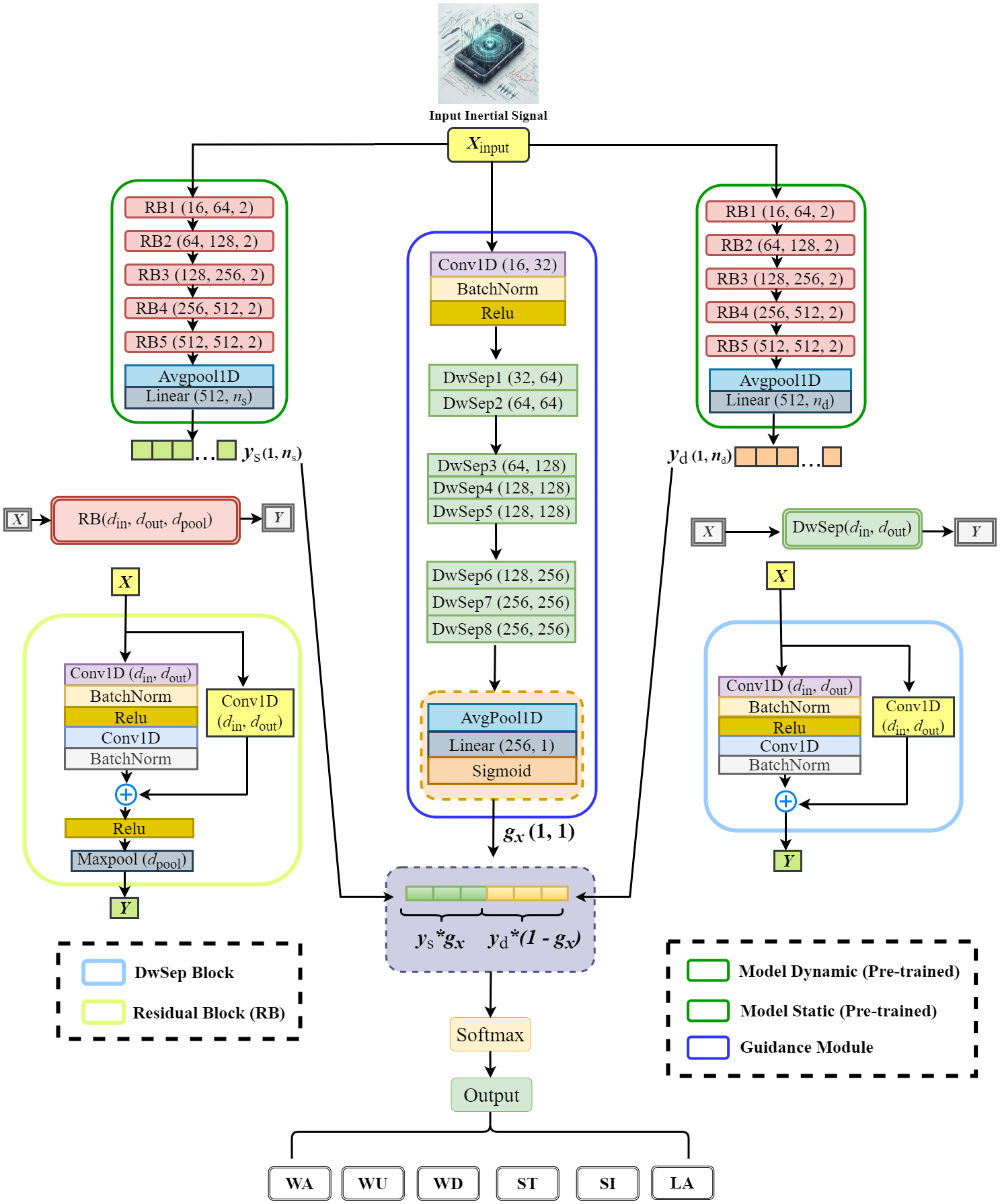}
    \caption{Detailed architecture of FusionActNet. The input signals are represented by $\mathbfit{X}_\text{input}$. The output of the model is one of the
    six activities [walking (WA), walking upstairs (WU), walking downstairs (WD), sitting (SI), standing (ST), and laying (LA).]}
    \label{fig:propose}
\end{figure}

At the next stage, a data sample $\mathbfit{X}_k$, which might be either from a static or dynamic superclass, is passed through both trained models, and the generated prediction vectors can be written as,
$$    \boldsymbol{y}_\text{s} = M_\text{s}(\mathbfit{X}_k); ~~~~
    \boldsymbol{y}_\text{d} = M_\text{d}(\mathbfit{X}_k)$$

Let $\boldsymbol{y}_\text{s}$ symbolize the prediction vector of the static model and $\boldsymbol{y}_\text{d}$ represent the prediction vector of the dynamic model for the signal $\mathbfit{X}_k$ in this context. It is important to note that due to the distinct attributes of the models $M_\text{s}$ and $M_\text{d}$, usually only one of the prediction vectors $\boldsymbol{y}_\text{s}$ or $\boldsymbol{y}_\text{d}$ would be relevant for the analytical objectives. However, both prediction vectors are included in the analysis at this point, as the final assignment of the sample $\mathbfit{X}_k$ to a superclass is still unknown. In order to get the desired output, a guidance factor $g_x$ is needed, which is obtained by processing the input $\mathbfit{X}_k$ with the guidance module as shown in Fig. \ref{fig:propose2}. Then $g_x$ is combined with previously obtained $\boldsymbol{y}_\text{s}$ and $\boldsymbol{y}_\text{d}$ to calculate the output activity label probabilities formulated as,
$$
\boldsymbol{y}_\text{pred} =g_x\cdot \boldsymbol{y}_\text{s} + (1 - g_x)\cdot \boldsymbol{y}_\text{d}
$$
with a loss function
\begin{equation}
\label{eqn:loss}
    L = \sum_{k=1}^{n}\text{Loss}(\boldsymbol{y}_{\text{pred},k},\boldsymbol{y}_{k})
\end{equation}
Here $\boldsymbol{y}_{k}$ is the true activity class.
The goal is to maximize the model's accuracy and dependability by minimizing the loss function, \textit{L} mentioned in Eq. (\ref{eqn:loss}).

\subsection{Network Architecture}
\label{ssec:net_arch}
In this study, an effective scheme has been proposed, that utilizes a guidance module for performing weighted concatenation of outputs from two expert models $M_\text{s}$ and $M_\text{d}$, to recognize static and dynamic human activities respectively. Because of performing a fusion operation considering the static/dynamic network outputs to recognize human activities from different sensor data, this network is named FusionActNet. The proposed model leverages the potential of 1D residual blocks as well as depth-wise separable CNN blocks, offering a dynamic and static pathway for information processing.
\subsubsection{Static Pathway $M_\text{s}$}
\label{sssec:subsubhead}
The static component of the FusionActNet is used for identifying the static human activities (e.g. sitting, laying, and standing) that comprise a series of residual blocks, as shown in Fig. \ref{fig:propose}, each of which consists of a convolutional layer, batch normalization, and rectified linear unit (ReLU) activation functions as well as a skip connection. Each residual block is represented by $\text{RB}(d_\text{in}, d_\text{out}, d_\text{pool})$ where $d_\text{in}$ and $d_\text{out}$ refer to the input and output dimensions of 1D convolution layers and $d_\text{pool}$ refers to the kernel size of the 1D max-pooling layer. The residual block begins with a convolutional layer with a 3x3 kernel and output channels $d_\text{out}$, followed by batch normalization and ReLU activation. Subsequently, another 3x3 convolutional layer with $d_\text{out}$ output channels is applied, followed by batch normalization. An identity convolutional layer with a 1x1 kernel is used to match dimensions for residual connections. A max-pooling layer with a down-sampling factor $d_\text{pool}$ is applied next. The final layer of the static pathway includes average pooling and a fully connected layer to reduce the output to $n_\text{s}$ output features, where, $n_\text{s}$ indicates the number of static activity labels.

\subsubsection{Dynamic Pathway $M_\text{d}$}
\label{sssec:subsubhead1}
The dynamic pathway is structurally identical to the static pathway but is expected to capture dynamic patterns or temporal information in the input dynamic data. It follows the same residual block structure but is trained with dynamic human activity data such as walking, walking upstairs, and walking downstairs,  having an equal number of output features, to the number of dynamic activity labels $n_\text{d}$ in the dataset. 

\subsubsection{Guidance Module}
\label{sssec:subsubhead12}
The neural module at the center of Fig. \ref{fig:propose} is called the guidance module, followed by a series of depthwise separable convolution blocks, DwSep, that consists of a 1D convolution layer, residual skip connection, the batch normalization, and ReLU activation. A 3x3 depth-wise convolution is applied for model size reduction and improving efficiency, followed by batch normalization and ReLU activation. A 1x1 point-wise convolution is used for dimensionality reduction, followed by batch normalization.
After traversing the series of DwSep blocks, the input proceeds through three key operations: average pooling, a linear layer, and a sigmoid block. Following the execution of these operations, a guidance factor denoted as $g_x$, is generated. The dimensionality of $g_x$ is (1,1) as the factor is prioritizing the output labels of a specific superclass, either static or dynamic. In other terms, the guidance factor can function as a classifier to discern whether the input data pertains to a static or dynamic context.

Upon passage through class-specific pre-trained static and dynamic models, the resultant outputs  $\boldsymbol{y}_\text{s}$ and $\boldsymbol{y}_\text{d}$ are consolidated through concatenation. Subsequently, following the generation of the guidance factor $g_x$, a weighted concatenation is executed on the previously merged output of $\boldsymbol{y}_\text{s}$ and $\boldsymbol{y}_\text{d}$. This strategic approach is employed to accord priority to the output of the pre-trained expert model. In simple terms, during the cases of static activity classes, the guidance factor $g_x$ will assign higher priority to the output of $M_\text{s}$. Similarly, the outputs of  $M_\text{d}$ will be multiplied by larger weights in the cases of dynamic activity classes, whereas the prediction of the other model will be assigned a very small weight.

\section{Results and Discussions}
\label{sec:res}
\begin{table}[h]
\caption{Evaluation results of different methods on UCI-HAR \cite{anguita2013public} dataset}
\label{uci_res}
\centering
\def\arraystretch{1.25}
\resizebox{0.45\textwidth}{!}{
    \begin{tabular}{lcccc}
    \hline
    \textbf{Method} & \textbf{Accuracy} & \textbf{Precision}  & \textbf{Recall} & \textbf{F1-score} \\ 
    \hline
    ResNet \cite{ferrari2019hand} & 0.9073 & 0.9095 & 0.8998 & 0.9046  \\
    Res-BiLSTM \cite{ullah2019stacked} & 0.9160 & 0.9150 & - & - \\
    Bi-LSTM \cite{mutegeki2020cnn} & 0.9270 & 0.9270 & - & - \\ 
    CNN \cite{wan2020deep} &  0.9271 & 0.9321 & 0.9321 & 0.9293  \\ 
    HAR-CT: CNN \cite{jaberi2022human} & 0.9406 & 0.9358 &  0.9359 &  0.9359  \\ 
    iSPLInception \cite{ronald2021isplinception} & 0.9510 & 0.9500 & - & - \\ 
    GRU+Attention \cite{wang2023novel} & 0.9600 & 0.9580 & - & - \\
    CNN-DCT \cite{zhao2018deep} & 0.9710 & - & - & - \\
    ConvBiLSTM-GRU \cite{sain2023human} & 0.9723 & - & - & - \\
    DL + SoTA \cite{challa2023optimized} & 0.9711 & - & - & 0.9714\\
    \textbf{FusionActNet}\\ \textbf{(Proposed)} & \textbf{0.9735} & \textbf{0.9700} &  \textbf{0.9710} &  \textbf{0.9739}  \\
    \hline
    
    \end{tabular}
} 
\end{table}

\begin{table}[h]
\caption{Evaluation results of different methods on MotionSense \cite{malekzadeh2018protecting} dataset}
\label{motion_res}
\centering
\def\arraystretch{1.25}
\resizebox{0.45\textwidth}{!}{
    \begin{tabular}{lcccc}
    \hline
    \textbf{Method} & \textbf{Accuracy} & \textbf{Precision}  & \textbf{Recall} & \textbf{F1-score}  \\ 
    \hline
    MFCC + SVM \cite{batool2019sensors} & 0.8935 & 0.8993 & 0.8968 &  0.8980  \\ 
    DT + BGWO \cite{jalal2020stochastic} &  0.9271 & 0.9321 & 0.9321 & 0.9293  \\ 
    Self-supervised TPN \cite{saeed2019multi} & 0.8901 & 0.8901 & 0.8899 & 0.8899  \\ 
    \textbf{FusionActNet}\\ \textbf{(Proposed)} & \textbf{0.9535} & \textbf{0.9500} &  \textbf{0.9507} &  \textbf{0.9521}  \\
    \hline
    \end{tabular}
} 
\end{table}
The performance of the proposed method is evaluated with four conventional evaluation criteria, namely accuracy, precision, recall, and F1-score, and results are discussed in this section.
For the validation of our proposed method, two benchmark datasets: UCI-HAR and MotionSense are utilized. A subject-based split, allocating 70\% of the data for training and validation and 30\% for testing is employed to ensure subject-independent validation. Specifically, 21 out of 30 subjects for UCI-HAR and 16 out of 24 subjects for MotionSense are used for training and validation, with the remaining subjects reserved for testing.
The proposed method achieved training and validation accuracies of 98.65\% and 98.10\% on the UCI-HAR dataset, and 97.15\% and 96.40\% on the MotionSense dataset, respectively. The training times are approximately 12 minutes for UCI-HAR and 20 minutes for MotionSense, conducted on an Nvidia P100 GPU. The hyperparameters are selected through extensive experimentation and validation of our datasets. Batch size is balanced for training speed and stability, a Reduce on Plateau strategy is implemented for adaptive learning rates, the Adam optimizer is chosen for its efficiency in handling sparse gradients, and Cross Entropy Loss is implemented in both stages of training for its suitability in classification tasks.
For the purpose of performance comparison, some state-of-the-art methods are considered and results are presented along with their corresponding evaluation metrics in Table \ref{uci_res} for the UCI-HAR dataset and Table \ref{motion_res} for the MotionSense dataset. 
The proposed method attained superior test accuracies of 97.35\% on the UCI-HAR dataset and 95.35\% on the MotionSense dataset, outperforming the existing state-of-the-art methods. The consistent performance across both benchmark datasets validates the stability of the proposed method's results. In order to present the performance in each class, the confusion matrixes of the corresponding datasets are shown in Table \ref{Tab:conf}. 

\begin{table}[ht]
\caption{Confusion matrix for UCI-HAR (Upper) and MotionSense (Below) datasets}
\centering
\label{Tab:conf}
\scalebox{0.85}{
\begin{tabular}{|c|c|c|c|c|c|c|} \hline 
\backslashbox{Actual}{Predicted} & WA & WU & WD & SI & ST & LA \\ \hline
WA & 0.95 & 0 & 0.05 & 0 & 0 & 0\\ \hline
WU & 0.01 & 0.98 & 0.01 & 0 & 0 & 0\\ \hline
WD & 0 & 0 & 0.93 & 0 & 0.07 & 0\\ \hline
SI & 0 & 0 & 0 & 0.94 & 0.06 & 0\\ \hline
ST & 0 & 0 & 0 & 0.02 & 0.98 & 0\\ \hline
LA & 0 & 0 & 0 & 0 & 0 & 1.00\\ \hline
\end{tabular}}
\scalebox{0.85}{
\begin{tabular}{|c|c|c|c|c|c|c|} \hline 
\backslashbox{Actual}{Predicted} & SI & ST & WD & WU & JG & WA \\ \hline
SI & 1.00 & 0 & 0 & 0 & 0 & 0\\ \hline
ST & 0 & 1.00 & 0 & 0 & 0 & 0\\ \hline
WD & 0 & 0 & 0.82 & 0.07 & 0.07 & 0.04\\ \hline
WU & 0 & 0 & 0.05 & 0.86 & 0.07 & 0.02\\ \hline
JG & 0 & 0 & 0.02 & 0.02 & 0.95 & 0.01\\ \hline
WA & 0 & 0 & 0.03 & 0.02 & 0.03 & 0.92\\ \hline
\end{tabular}
}
\end{table}
The main advantage of our proposed method lies in its utilization of two dedicated models for capturing static and dynamic behaviors, coupled with a guidance mechanism for refining final predictions. By employing this two-stage approach, our proposed scheme enhances performance by incorporating complementary information from different models and fine-tuning predictions through the guidance mechanism. However, a limitation of our method is the slightly higher training time due to the two-stage training process. Another limitation of the proposed method is the misclassification of closely similar activities like 'walking upstairs' or 'walking downstairs' in the 'Motionsense' dataset shown in Table \ref{Tab:conf}.
\section{Conclusion}
\label{sec:con}
This letter introduces an effective scheme for human activity recognition, incorporating two distinct approaches within a comprehensive multi-structural framework referred to as FusionActNet employing residual networks and a guidance module with depth-wise separable CNN blocks. For effectively classifying static and dynamic activities, two distinct models are employed to capture the underlying pattern of closely related activities, unlike existing works. Additionally, the guidance module is included for emphasizing the final prediction vectors through weighted concatenation, which distinguishes this work from existing training processes. The results obtained using the proposed method are validated using only sensor-based data, suggesting that in future research, efforts could be made to generalize the approach for all forms of data used in activity recognition. Despite a few limitations, the proposed FusionActNet achieves state-of-the-art accuracy on both UCI-HAR and Motion-Sense datasets, demonstrating stability in handling data overlap scenarios that showcase promising advancements in activity recognition techniques, with potential applications in various domains requiring precise human activity classification.\\
\bibliographystyle{IEEEbib}
\bibliography{strings}

\end{document}